# On The Robustness of a Neural Network

El Mahdi El Mhamdi
elmahdi.elmhamdi@epfl.ch

Rachid Guerraoui
rachid.guerraoui@epfl.ch

Sébastien Rouault
sebastien.rouault@epfl.ch

Swiss Federal Institue of Technology, Lausanne, Switzerland

*Abstract*—With the development of neural networks based machine learning and their usage in mission critical applications, voices are rising against the *black box* aspect of neural networks as it becomes crucial to understand their limits and capabilities. With the rise of neuromorphic hardware, it is even more critical to understand how a neural network, as a distributed system, tolerates the failures of its computing nodes, neurons, and its communication channels, synapses. Experimentally assessing the robustness of neural networks involves the quixotic venture of testing all the possible failures, on all the possible inputs, which ultimately hits a combinatorial explosion for the first, and the impossibility to gather all the possible inputs for the second.

In this paper, we prove an upper bound on the expected error of the output when a subset of neurons crashes. This bound involves dependencies on the network parameters that can be seen as being too pessimistic in the average case. It involves a polynomial dependency on the Lipschitz coefficient of the neurons' activation function, and an exponential dependency on the depth of the layer where a failure occurs. We back up our theoretical results with experiments illustrating the extent to which our prediction matches the dependencies between the network parameters and robustness. Our results show that the robustness of neural networks to the average crash can be estimated without the need to neither test the network on all failure configurations, nor access the training set used to train the network, both of which are practically impossible requirements.

## I. INTRODUCTION

Beating the world champion in the game Go [1], recognizing images with supra-human precision [2] or, more recently, diagnosing skin cancer with dermatologist level precision [3], all of the impressive progress made in machine learning and artificial intelligence in the past decade is due to the use of neural networks [4], [1], [5]. Initially inspired by the working principles of the natural nervous system [6], a neural network is a mathematical object with outstanding, yet not very well understood computational capabilities [7].

Today, most of the implementations view a neural network as a software, a mathematical abstraction that is only simulated on top of Turing machines [8]. This induces a computational bottleneck [9], due to the transition from a discrete digital computation to a continuous, analogue computation, each time the software (simulating a neural network) queries the hardware and gets the result back.

One of the paths to scale up machine learning is to use a new kind of hardware that does not suffer from the aforementioned bottleneck. Recent progress in neuromorphic hardware brings such a solution, going beyond the Von Neumann paradigm (relying on logical circuits) and using electronic chips that are themselves neural networks [10], [11], [9], [12]. A year ago, teams from IBM reported [13], [14] a successful neuromorphic implementation of neural networks that requires a running power as low as 25 mW to 275 mW to perform image recognition tasks that would require orders of magnitude more energy on a classical computer.

Neural networks are now considered for critical and safety-sensitive applications such as flight control [15], radars [16] or self-driving cars [17]. If one cannot yet elucidate all of the working principles of neural networks, one should at least guarantee their robustness to failures in order to use them safely.

To achieve robustness, one can stick to the view of the entire neural network as a single piece of software [18], replicate this piece on several machines, and use classical state machine replication schemes to enforce the consistency of the replicas [19]. In this context, no neuron is supposed to fail independently: the unit of failure is the entire machine hosting the network. This coarse-grained approach is the one taken by recent works on the robustness of machine learning [20], [21], [22]. However, forcing an entire network to run on a single machine clearly hampers scalability as stated above. One could also consider strict subsets of the neural network as different pieces of software, each running on one Turing machine [23]. In this case, classical replication schemes can still be applied, but one has to face usual distributed computing problems, e.g., to handle the synchronicity of messages exchanged between subsets of the network [24], and even-though subsets of the network are distributed over machines, the computation still goes through the Von Neumann bottleneck described above: we are still using Boolean circuits to simulate the analogue and continuous computation of the neural network.

The scalability promise of neuromorphic hardware, calls for going one step further and considering each neuron as a *single* physical entity that can fail *independently*, i.e., to go for genuinely distributed neural networks [10]. In this setting, the unit of failure is one single neuron or synapse, and not a whole machine.

In fact, the enthusiasm around the efficiency of neuromorphic hardware is not a futuristic hype. This type of hardware is *already* used in sensitive applications for which robustness to the failures of single neurons is crucial. As of January 2017, U.S Air Force acknowledged the use of an IBM *Brain-inspired* chip that uses 20 to 30 times less power than a classical computer for military applications. Unsurprisingly, this chip is made of neuromorphic circuits [25].

Now say, for the sake of example, that we would like to send a neuromorphic chip made of 150 neurons (a rather small amount) in a satellite, and we want it to run for (at least) 10 years. Let us suppose that, due to solar radiations, we expect a maximum loss of 50 neurons over those 10 years. Evidently,

*losing*[1] neurons will change the chip output for every input by some $\epsilon(input)$, and if $\epsilon(input)$ gets too *large* for some inputs, the chip might become useless/dangerous according to its purpose. So, prior to asserting the chip usefulness/safety, one must obtain $\epsilon(input)$ for every possible input, or at least bound them with $\max_{inputs} \epsilon(input)$. In the absence of a theoretical guarantee, we will have to test our chip for every possible combination of 50 neurons lost among 150 and every possible input. In our example, this represents $\binom{150}{50} \approx 2 \times 10^{40}$ different neural networks, that must then be tested for every possible input. And if this safety assessment is performed for the aforementioned IBM chip, containing a million neurons, the test will require more executions than there are atoms in the observable universe ($10^{80}$). This motivates the need for a theoretically–grounded robustness guarantee that does not require unrealistic testing.

Traditionally, neural networks were considered robust in the sense that the failure of neurons gracefully degrades their accuracy [8] and that this degradation can be compensated by additional learning phases [26]. To the best of our knowledge, the work on failures at the level of individual neurons has been mostly experimental [27], [28], [29], [30], [31], or, when it included a theoretical analysis, it focused on bounding the effect of the lost of a single neuron [32] or the cost in terms of additional learning [26].

It is actually well known that the failure of neurons can be tolerated through additional learning phases [8]. Nevertheless, stopping a neural network and launching additional learning phases is simply unimaginable for critical real-time applications. One can also consider specific learning schemes a priori that make it possible to tolerate failures a posteriori, e.g., shutting down parts of the network while learning, in order to cope with failures at run-time (dropout) [27], [33]. Yet, no theoretical result exists to predict the effect of dropout on robustness. The question we address can be actually stated as follows: if (a) we do not make any assumption on the learning scheme and (b) we preclude the possibility of adding learning phases after computations have started, what is the maximum number of faulty neurons that can be tolerated by a neural network?

As we pointed out in [34], the answer to this question is simply *none*. Indeed, why would a neural network tolerate failures if it was not specifically devised with that purpose in mind? More specifically, if the failure of a number of neurons does not impact the overall result, then these neurons could have been be eliminated from the design of that network in the first place. In fact, the reason why the question is nontrivial is *over-provisioning* [35]. Neural networks are rarely built with the minimal number of neurons to perform a computation. To estimate exactly this minimal number, one needs to know the target function the network should approximate, which by definition is unknown[2]. In fact, it has been experimentally observed that over-provisioning [8], [27] leads to robustness. In a recent work [34], [36], we provided a first theoretical guarantee of neural networks robustness, given an over-provision budget. However, we focused on worst case situations that can be too pessimistic compared to the average case of practical ones.

In this paper, we prove that a practical estimator of neural networks robustness can be computed directly from the network parameters, without the costly comparison on all possible inputs and all crash situations, and that the average case suffers from similar dependencies as the worse case. We validate experimentally our equations and show how the predicted dependencies on each parameter are reflected in practice. We do that both on randomly generated neural network of small size (for which testing all the possible crash situations is possible), and on realistic neural networks that recognize handwritten digits (on which we tested up to a certain value of number of failures).

We establish this relation and provide experimental evidence of the role played by each of the key parameters of a neural network. To do so, we consider the classical and general model of a multilayer perceptron [8], which is the most general model to mathematically abstract the topologies used in modern deep learning such as the popular convolutional scheme [23], [33].

We experimentally explore this predictive method on the average case for realistic tasks such as image recognition. To do so, we introduce two quantities: $\Omega$ that can be computed using all possible inputs and all possible crash situations (expensive computation), and $Erf$, that can only be computed by some constants of the network (cheap computation). We experimentally show how $\Omega$ varies when we vary key constants of the network, then we provide experimental evidence that $Erf$ is a safe estimator of $\Omega$.

Our experiments show that the value of the Lipschitz coefficient of the activation function, together with the distribution of large synaptic weights and the depth of the network are the key parameters to control how errors propagate in a neural network. In short, robust neural networks are those that (1) have low and evenly distributed synaptic weights, (2) are shallow if the Lipschitz coefficient is smaller than 1, and deep if this coefficient is larger than 1, (3) when depth is a requirement, can use an activation function with a low Lipschitz coefficient. If choosing such an activation function is not possible (in practice, not all activation functions enable networks to learn with the same efficiency [2]), alternatives are functions for which the Lipschitz behavior is restricted to a narrow area.

Finally and most importantly, our work shows how an easily computable quantity can be used to estimate the robustness of a network without the costly testing that requires as much computations as there are possible inputs to the networks and combinatorial possibilities of sub-networks when we consider all possible crashes. Interested readers can find both the source code and the raw data of our experiments in the following repository [37].

The rest of the paper is organized as follows. Section 2 gives a general description of a neural network of which components can fail and the criteria we follow to assess robustness. Section 3 provides the theoretical framework for our robustness estimator and proves the dependencies between robustness and networks parameters in the average case. Section 4 describes our main experimental results. Section 5 explores the main trade-offs between robustness and ease of learning. Section 6 summarizes the paper and discusses related work.

---

[1] What *losing* a neuron means is specified in section II-B, definition 2.

[2] In machine learning, we only know a finite number of the values of the target function: the training set

## II. MODEL OF COMPUTATION

### A. Main Components

We use the same model as in [34]. We view a neural network as a distributed system comprised of computing nodes, *neurons*, and communicating channels, *synapses*:

**Processes:** *Neurons* are the computing nodes: we assume them to be unreliable in the sense that they can stop computing, a situation we will call a *crash*. The failure of any node is independent from the failure of any other node. Neurons communicate via message-passing [24] through directed communication channels called *synapses*. During the operating phase, the communication is always from the side of the input to that of the output[3]. In the experiments, we will say that a neuron was *killed* when we purposely switch it off, the network will behave as if the killed neuron has crashed.

**Channels:** *Synapses* have a similar reliability model to the neurons with two situations: either they are correct, or they crash and do not transmit the messages. The failure of a synapse is also independent from that of other synapses and neurons. Based on [34], synapses failures can be abstracted as mathematically equivalent to some related neurons' failures, our experimental work focuses therefore on neurons failures.

A notable property of synapses is that they are *weighted*, each weight represents the importance that the neuron on the input side of the synapse represents for the neuron on the output side. To build a network in a supervised fashion, one uses a training set of inputs for which correct outputs are provided. Typically, the training set of a network is comprised of thousands of labelled files. Although the theory behind learning procedures is out of the scope of this paper, one should keep in mind that these procedures consists of looping through the training set for enough time until the convergence of the values of the weights to some ideal point that corresponds to the computation model described in what follows.

**Network:** A *neural network* is a distributed system comprised of neurons, connected by synapses, in this paper, we consider the model of *feed-forward* neural network. A simple yet general model that encompasses today's most popular models such as the convolutional network used in deep learning [4]. Feed-forward network are organized in *layers*, in this paper we use the conventional notation [8] and refer to the number of layer by $L$, to the number of neurons in the $l^{th}$ layer by $N_l$, and to the weight of the synapse linking the $j^{th}$ neuron in the $(l-1)^{th}$ layer to the $i^{th}$ neuron in the $l^{th}$ layer by $w_{ij}^{(l)}$. Figure 1 provides a simple illustration of this notation.

**Computation:** Let $\epsilon$ be any positive real number (an accuracy level), and $F$ any continuous function mapping $[0,1]^d$ to $[0,1]$. A neural network implements a function $F_{neu}$ as given by Equation 1 and Figure 1, such that $F_{neu}$ approximates the target function $F$ with an accuracy $\epsilon$ in a series of local computations consisting of linear combinations of the outputs from the layer on a neuron's input side (Figure) -with the synaptic weights $w_{ji}^{(l)}$ as coefficients in that linear combination- to which neurons apply the activation function $\varphi$ after adding a bias $b_j^{(l)}$, a monotone and K-Lipschitz-continuous non linearity. We recall that a K-Lipschitz-continuous function $\varphi$ is a function such that[4] $|\varphi(y) - \varphi(x)| \leq K.|y - x|$ for every inputs $x$ and $y$. In this paper, we consider two classes of such activation functions: (a) the bounded sigmoid function, and (b) the unbounded linear rectifier (Figure 2).

$$F_{neu}(\mathbf{X}) = \sum_{i=1}^{N_L} w_i^{(L+1)} y_i^{(L)} \quad (1)$$

$$\text{with } y_j^{(l)} = \begin{cases} x_j & \text{for } l = 0 \\ \varphi(s_j^{(l)}) & \text{for } 1 \leq l \leq L \end{cases} \quad (2)$$

$$\text{and } s_j^{(l)} = \sum_{i=1}^{N_{l-1}} (w_{ji}^{(l)} y_i^{(l-1)} + b_j^{(l)}) \quad (3)$$

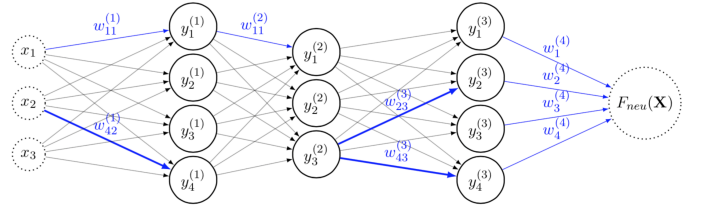

Fig. 1. A (feed forward) neural network (solid nodes and edges), with $d = 3$, $L = 3$, $N_2 = 3$ and $N_1 = N_3 = 4$. Input and output nodes (dotted) are not considered as parts of the network, but as its clients. For readability, only some synaptic weights are represented (bold blue). $\mathbf{X} = (x_1, \ldots, x_d)$.

The effectiveness of feed-forward neural networks relies on a fundamental theorem [38] that guarantees their universal approximating power with as few[5] as one single layer.

**Definition 1.** *(Approximation) We denote by $A = C([0,1]^d, [0,1])$ the space of continuous functions mapping $[0,1]^d$ to $[0,1]$, and by $\|.\|$ an appropriate norm on A. $F_{neu}$ as defined by Equation 1 is said to be a neural $\epsilon$-approximation of a target function $F \in A$ for the norm $\|.\|$ if we have: $\|F - F_{neu}\| \leq \epsilon$.*

---

[3]During the learning phase, there are also communications in the other direction but this paper does not tackle learning

[4]For differential functions, $K$ is simply the maximal value of the first derivative (steepest slope).

[5]Note that universality for L=1 is harder to obtain than for $L > 1$: fewer layers to approximate the target function.

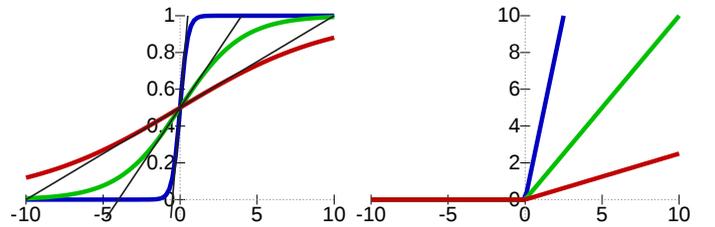

Fig. 2. (Left) The profile of a sigmoid function, centered around 0 and tuned with several values of K. The larger is K, the steeper is the slope and the more discriminating is the activation function at each neuron. The *Lipschitz behavior* in the text means the attitude of sigmoid in the area given by the tangent on 0, as we see, the smaller is K the larger is this area. (Right) The profile of a linear rectifier (ReLu) with similar variation of the Lipschitz coefficient. The Lipschitz behavior for ReLu concerns all positive values at the input.

Naturally, for the worst case analysis, one might consider $\|.\|$ as the infinite norm, while for an average case analysis the $L^1$ norm will be more appropriate.

**Universality.** We recall the universality theorem for a single layer network[6]: Let $d$ be any integer and $\varphi : \mathbb{R} \to [0, 1]$ a strictly-increasing continuous function, such that $\lim_{x \to -\infty} \varphi(x) = 0$ and $\lim_{x \to +\infty} \varphi(x) = 1$. Given any function $F \in A$ and $\epsilon > 0$, there exist an integer $N(\epsilon)$, and a set of coefficients $(w_{ji}^{(1)})_{1 \leq j \leq N(\epsilon)}^{1 \leq i \leq d}$ and $(w_i^{(2)})_{1 \leq i \leq N(\epsilon)}$ such that $F_{neu}$ defined in Equation 1 is a neural $\epsilon$-approximation of $F$.

*B. Evaluating Robustness*

Given an over-provisioned network implementing $F_{neu}$ and achieving $\epsilon'$-acccuracy ($\epsilon' \leq \epsilon$), we would like to evaluate how many neurons can crash without harming the $\epsilon$-approximation of $F$ by $F_{neu}$. We use the same definitions as in [34].

**Definition 2.** *(Failures) We say that a neuron $i$ in layer $l$ crashes when neuron $i$ stops sending values, in which case $y_i^{(l)}$ is considered[7] to be equal to 0 by other neurons[8].*

**Definition 3.** *(Expected Robustness) We say that a neural $\epsilon$-approximation $F_{neu}$ of a target function $F$ realized by $N$ neurons tolerates $N_{fail}$ crashed neurons on average, if we have $\mathbb{E}_{|I_{fail}|=N_{fail}} \|F - F_{fail}\| \leq \epsilon$, where the expectation runs on all subsets of neurons $I_{fail} \subset \{1, \cdots, N\}$ of size $N_{fail}$, and $F_{fail}$ is a modified version $F_{neu}$ for the neurons in $I_{fail}$ according to Definition 2.*

In the case where the activation function is bounded, worst-case theoretical analysis reveals [36][9] that the robustness of the network is completely encompassed in the weight distribution and the network topology: a network of $L$ layers with a $K$-Lipschitz activation function tolerates the crash of $f_l$ neurons per layer $l$ if the quantity $Erf$ defined in Equation 4 is smaller than the error margin $\epsilon - \epsilon'$ enabled by the over-provision.

$$Erf = \sum_{l=1}^{L} \left( C_l f_l K^{L-l} w_m^{(L+1)} \prod_{l'=l+1}^{L} (N_{l'} - f_{l'}) w_m^{(l')} \right) \quad (4)$$

In Equation 4, $w_m^{(l)} = max(|w_{ji}^{(l)}|, (j,i) \in [1, N_l][1, N_{l-1}])$ is the maximum norm of the weights of the incoming synapses to layer $l$, $K$ is the Lipschitz coefficient of the activation function, $N_l$ the total number of neurons in layer $l$, $f_l$ the total number of crashed neurons in layer $l$ and $C_l$ is the maximum value on the neurons output in layer $l$ (1 for sigmoid, and network-specific for ReLu).

Taking the average weight in Equation 4, we generalize the formula on $Erf$ to (1) the average case which we experimentally show to be less pessimistic than the worst case (especially for the saturating activation function sigmoid), and to (2) the case where the activation function is not bounded, which is more relevant in today's applications where the linear rectifier (ReLu) is becoming more popular [2] than the bounded sigmoid function. Then we test the predictive power of our formula by computing the averages and worst cases on all the available inputs for a given network.

We proceed by generating random networks, assigning values to synaptic weights that follow a normal distribution around 1 with a standard deviation of 5. It is natural to consider each of those networks as an approximation of itself with $\epsilon' = 0$, then we take the error made by the network - after some neurons are purposely crashed - as an estimator of $\epsilon$.

To provide more realistic orders of magnitude, we train feed forward neural networks with different activation functions (sigmoid, and the linear rectifier) and with different values of the Lipschitz coefficient on the standardized dataset MNIST of handwritten digits [39], and test them against neurons removal.

In the following, we introduce a function $\Omega$ such that for a given input $\mathbf{X}$, $\Omega(\mathbf{X}) = \|F_{neu}(\mathbf{X}) - F_{alt}(\mathbf{X})\|$ where $F_{alt}$ is an altered version of the network due to failures. We say that a network *tolerates $f$ failures while keeping the error smaller than $\epsilon$* on input $\mathbf{X}$, if for every altered version of the network missing $f$ neurons we have $\Omega(\mathbf{X}) < \epsilon$.

For a given number of failures $f$, $\Omega_{av}$ (for "average") $\Omega_{mav}$ (for "max average") and $\Omega_{max}$ are respectively the average of $\Omega$ on all possible combinations of $f$ failures and all inputs, the average on all possible inputs of the maximal $\Omega$ given by $f$ failures, and finally the maximum on all possible inputs of the maximal $\Omega$ given by $f$ failures.

Similarly, we introduce variants of $Erf$ as defined in Equation 4 and note them $Erf_{max}$, $Erf_{av}$ (for average). Note that there is no $Erf_{mav}$ (for max average) since $Erf$ is input-independent.

One should keep in mind that computing $Erf$ and its variants **only requires access to the network properties** (weights and topology), while computing $\Omega$ requires the same amounts of computation needed for $Erf$ multiplied by the number of possible inputs the network can work on and by the combinatorial explosion due to all the sub-networks possible when $f$ neurons are crashed.

While $\Omega$ gives a precise evaluation of the robustness of a network, our goal is to show how $Erf$, as cheap to compute as it could be, is still a fair estimator of $\Omega$ and thus of the robustness of the network.

### III. THEORETICAL ESTIMATOR

In this section, we extend the results of [34] to derive a bound on the average error made by the network, when neurons are crashed at random. In all of the following, $\|.\|$ refers to the $L^1$ norm on real valued functions, recall that for two functions $f$ and $g$, $\|f - g\| = \int_{x \in H} |f(x) - g(x)| dx$, where $H$ is the input space of the functions $f$ and $g$.

To understand the rationale behind our robustness estimator $Erf$, let us start by exploring what happens on average in a single layered network when random sets of $N_{fail}$ neurons crashe. We later extend the bound for a multi-layer network.

---

[6]The interested reader can refer to the proof of [38].
[7]The strictly-increasing activation function $\varphi$ does not allow a correct neuron to output value 0.
[8]Remember that we assume synchronous transmission.
[9]This is a companion theoretical technical report that proves Equation 4 in the worst case situation.

**Theorem 1.** *Let $F$ be any function mapping $[0,1]^d$ to $[0,1]$. Let $\epsilon$ and $\epsilon'$ be any two positive real numbers such that $0 < \epsilon' \leq \epsilon$. For any neural $\epsilon'$-approximation $F_{neu}$ of $F$ (Definition 1) and any integer $N_{fail}$: If $N_{fail} \leq \frac{\epsilon - \epsilon'}{C \cdot w_{av}}$ where $C$ is the maximum absolute value of the activation function[10], $w_{av} = mean(|w_i^{(2)}|, i \in [1, N])$ is the mean absolute value of a weight from the single layer to the output node, then $F_{neu}$ is a neural $\epsilon$-approximation of $F$ that tolerates $N_{fail}$ crashed neurons on average (Definition 3).*

*Proof:* In all of this proof and for notational convenience, $\mathbb{E}v$ means the expectation of the random value $v$ over all possible random sets $I_{fail}$ containing $N_{fail}$ crashed neurons.

Let $\epsilon$ and $\epsilon'$ be any two positive real numbers such that $0 < \epsilon' \leq \epsilon$. Applying the universality theorem[11] on $F$ and $\epsilon'$, let $F_{neu}$ be a neural $\epsilon'$-approximation of $F$ with $N$ the number of neurons of $F_{neu}$ and $w_{av}$ the average absolute value of the weights from the single layer of $F_{neu}$ to the output.

Let $N_{fail}$ be any integer such that $N_{fail} \leq \frac{\epsilon-\epsilon'}{C \cdot w_{av}}$. Denote by $F_{fail}$ a random variable that corresponds to the modified values of the neural function $F_{neu}$ after $N_{fail}$ neurons crashed: $F_{fail} = \sum_{i=1, i \notin I_{fail}}^{N} w_i^{(2)} y_i$, where $I_{fail}$ is a set containing $N_{fail}$ crashed neurons.

By the triangle inequality and taking the expected value [12] we have:

$$\mathbb{E}\|F - F_{fail}\| \leq \mathbb{E}\|F - F_{neu}\| \\ + \mathbb{E}\|F_{neu} - F_{fail}\|. \quad (5)$$

Since $F_{neu}$ is not affected by the random crashed neurons (only $F_{fail}$ is) and since it is an $\epsilon'$-approximation of $F$ we have:

$$\mathbb{E}\|F - F_{neu}\| = \|F - F_{neu}\| \leq \epsilon'. \quad (6)$$

From the definition of $F_{fail}$, we have, for every input ($\mathbf{X}$): $F_{neu}(X) - F_{fail}(\mathbf{X}) = \sum_{i=1, i \in I_{fail}}^{N} w_i^{(2)} y_i(\mathbf{X})$.

The random nature of $F_{fail}$ is now fully represented by the values $\{w_i^{(2)}, i=1, i \in I_{fail}\}$ for which we know the expected value $w^{av}$.

Using the Jensen inequality we obtain:

$$\mathbb{E}\|F_{neu} - F_{fail}\| \leq \sum_{i=1, i \in I_{fail}}^{N} \mathbb{E}|w_i^{(2)}| \cdot \|y_i\| \quad (7)$$

---

[10]$C = 1$ for sigmoid, and equals the diameter of the input space times the Lipschitz coefficient in the case of ReLu. We can expect finer precision on $C$ if we average failure situations and take maximum value on inputs but this requires knowledge of the diameter of the input space.

[11]As in [34], the existence of a neural approximation for a given target function is taken here as granted by the universality theorem.

[12]When dealing with the worst case in [34], we applied the triangle inequality point-wise, here we are considering the $L^1$ norm as announced above, since it encompasses an averaging on all the input space, and we take expected value $\mathbb{E}$ that encompasses averaging on all the different sets of crashed neurons.

Note that if we impose on the weights to be all positive, and when the $y_i$'s are positive functions (the case for sigmoid and ReLu described in the model), Inequality 7 is an equality and the bounds on the expected values in this proofs are tight.

By definition of $w_{av}$ and the hypothesis on the activation function, $\mathbb{E}|w_i^{(2)}| = w_{av}$ and $y_i(\mathbf{X}) \leq C$ for all $\mathbf{X}$ and $i$. Inequality 7 becomes:

$$\mathbb{E}\|F_{neu} - F_{fail}\| \leq \sum_{i=1, i \in I_{fail}}^{N} C \cdot w_{av} = N_{fail} \cdot C \cdot w_{av} \quad (8)$$

Using inequalities 5, 6 and 8 we obtain: $\mathbb{E}\|F - F_{fail}\| \leq N_{fail} \cdot C \cdot w_{av} + \epsilon'$.

Since $N_{fail} \leq \frac{\epsilon - \epsilon'}{C \cdot w_{av}}$, we have $\mathbb{E}\|F - F_{fail}\| \leq \epsilon$.

With the last inequality, we proved that the condition on $N_{fail}$: $N_{fail} \leq \frac{\epsilon - \epsilon'}{C \cdot w_{av}}$, guarantees that $F_{neu}$ is robust according to Definition 3. ∎

In the previous proof, we used two key facts:

- $F_{neu}$ is an $\epsilon'$-approximation
- The effect of failures is bounded by $\epsilon - \epsilon'$

Then we plugged this into Inequality 5. The first fact will in fact always hold, thanks to the over-provision. All the work to guarantee robustness is in fact about assuring that the average deviation from the full neural network ($\mathbb{E}\|F_{neu} - F_{fail}\|$ is bounded by our error budget $\epsilon - \epsilon'$ (enabled by the over-provision).

Denote by $Erf_{av}$ the quantity given by replacing maximal weights in Equation 4 by average absolute value of weights, this quantity is smaller that $Erf$ and is therefore a finer upper bound. The following theorem guarantees that the expected deviation from the full neural network $\mathbb{E}\|F_{neu} - F_{fail}\|$ is bounded by $Erf_{av}$, therefore, it is sufficient to compare $Erf_{av}$ to the error budget to assess the average robustness.

**Theorem 2.** *Consider a neural network containing $L$ layers. If in each layer $l$, $f_l$ neurons among the $N_l$ neurons are crashed, then the effect on the output is bounded as follows:*

$$\mathbb{E}\|F_{neu} - F_{fail}\| \leq Erf_{av} \quad (9)$$

*where $F_{neu}$ is the nominal neural function, $F_{fail}$ a random variable describing the neural function whith crashes (as in the single layer formalism), and $w_{av}^{(l)} = mean(|w_{ji}^{(l)}|, (j,i) \in [1, N_l][1, N_{l-1}])$ is the mean absolute value of the weights of the incoming synapses to layer $l$.*

*Proof:* We proceed by induction on $L$ the number of layers.

**Initiation.** When $L = 1$, $Erf_{av} = C_1 \cdot f_1 \cdot w_{av}$ and the result follows from Inequality 8 considering $N_{fail} = f_1$ the number of crashed neurons in the single layer of the network.

**Induction step.** Assume Theorem 2 holds for networks with up to some number of layers $L \geq 1$. Consider now a network consisting of $(L+1)$ layers.

We proceed with the same methodology as in [34] that we adapt to the expected value of the error. The layered structure of the network enables us to see each of the $N_{L+1}$ neurons of the $(L+1)^{th}$ layer, first as an output to an $L$-layer network (all the nodes to the input side of that neuron), and second, after applying the activation function, as a neuron in a single-layer neural network (consisting of the $(L+1)^{th}$ layer alone).

In this last $(L+1)^{th}$ layer, we can distinguish two subsets of neurons:

1) (Crashed neurons at layer L+1) A subset of $f_{L+1}$ crashed neurons, that yields, as in the initiation step (sigle layer), an error of at most $f_{L+1} w_{av}^{(L+2)} C_{L+1}$.
2) (Correct neurons at layer L+1) A subset of $N_{L+1} - f_{L+1}$ correct neurons. Those neurons transmit to the output side (their right side in Figure 1), in addition to their nominal value, the error $E$ of the $L$-layer neural network on the output side of layer (L+1), multiplying it *on average* by at most the average synaptic weight from layer L to layer $(L+1)$, $w_{av}^{(L+1)}$, and the Lipschitz constant K, yielding an error with the following expected value $\mathbb{E}(E \cdot (N_{L+1} - f_{L+1})K)$.

By the induction hypothesis we have:

$$\mathbb{E}(E) \leq \sum_{l=1}^{L} C_l f_l K^{L-l} \prod_{l'=l+1}^{L+1} (N_{l'} - f'_l) w_{av}^{(l')}$$

As the output node is linear the errors mentioned in 1 and 2 are added, so are their expected values, which yields a total expected error bounded as follows (using Jensen inequality as in the proof of Theorem 1):

$$\mathbb{E}\|F_{neu} - F_{fail}\| \leq f_{L+1} w_{av}^{(L+2)} C_{L+1} + (N_{L+1} - f_{L+1}) K \mathbb{E}(E)$$
$$\leq \sum_{l=1}^{L+1} C_l f_l K^{L+1-l} \prod_{l'=l+1}^{L+2} (N_{l'} - f'_l) w_{av}^{(l')}$$

which is the desired bound for an $(L+1)$-layer network.

By induction, Theorem 2 is true for any integer $L \geq 1$. ∎

## IV. EXPERIMENTAL RESULTS

In subsections IV-A and IV-B, we present our experimental findings based on randomly generated 4-layers networks, each layer of which contain 4 neurons. We studied those networks with both sigmoid and ReLu activation functions, and they all have a linear mono-dimensional output.

In subsection IV-C, we report on a couple of more realistic neural networks, namely networks that we trained to recognize images from the standardized MNIST database of handwritten digits. One category of networks was constituted of 3-layers networks of 8 neurons each, using the ReLu activation function and the other category is constituted of single-layered networks of 48 neurons using the ReLu activation as well, and trained with the dropout procedure [27] that consists of randomly switching off neurons during the training phase. In both categories, the output is 10-dimensional, each component $i$ provides the probability that the given input is the digit $i$.

The dimensions of our networks can seem rather small if we forget the introductory note: looking at all crash situation induces a *combinatorial explosion*. The reader should keep in mind that each of the data points presented below is the result of averaging the results of as much runs as there are possible combinations of lost neurons and inputs. For instance, in figure 10, measuring $\Omega$ and its standard deviation for up to 5 crashed neurons among 48 took about two months on our computer. Indeed, there are 60000 data points in the MNIST dataset, which yields a total of $60000 \times \left(\binom{48}{1} + \binom{48}{2} + \binom{48}{3} + \binom{48}{4} + \binom{48}{5}\right) \approx 10^{11}$ different runs, and one run costs approximately $50\mu s$ with our sequential algorithm[13].

Interested readers can find both the source code and the raw data of our experiments in the following repository [37].

### A. Role of the Lipschitz coefficient and the depth

The activation function is behind the non-linearity of neural networks, and is therefore one of the key factors in the their computational expressivity. The theoretical analysis showed that, in a network with $L$ layers, the error made on the output when a neuron in layer $l$ crashes is proportional to $K^{L-l}$, where $K$ is the Lipschitz-coefficient (c.f Figure 2) of the activation function. In this subsection, we will explore the role played by the steepness of parameter on the robustness of a network, more precisely, on how this parameter amplifies or reduces the propagated error to the output.

On the role of the Lipschitz coefficient on robustness, our experiments confirms that: (a) the consequence (on the output error) of a crash is polynomial on the Lipschitz coefficient, (b) the degree of this polynomial dependency is, in the worst case, equal to the depth of the crash: how far is the crash from the output layer (Figure 5), thus an exponential dependency on the depth where the crash occurs.

In figures 3 and 4, we plot the evolution of the variables $\Omega$. The attitudes are representative of what we on the many generated networks and we observe the following:

1) While ReLu acts as a propagator of the polynomial dependency on K predicted by $Erf$, sigmoid has a saturating effect.

This can intuitively be understood from Figure 2, and the notion of *area of Lipschitz behavior*: for sigmoid, this area is restricted to a narrow interval that becomes narrower as K grows, while this area consists of all the positive inputs for ReLu, whatever is the Lipschitz coefficient.

2) In the case of ReLu as an activation function, estimating the propagated error by $\Omega_{max}$ is on average two orders of magnitude more pessimistic than considering $\Omega_{mav}$. The saturating effect of sigmoid makes of $\Omega_{mav}$ a good estimator of $\Omega_{max}$ (only $\Omega_{mav}$ is kept for plots for sigmoid for figures readability, interested readers can check the raw data provided in [37]). The difference between different $\Omega$s translates the activation function the effect of the size of the input space, as noted in the proof of Theorem 1 and footnote 10.

3) The slope of Figure 3 corresponds exactly to $K^4$ as given by Equation 4. In Figure 4 this is the case only when $K << 1$.

---

[13] A parallelization of the algorithm would not be worth the benefit: on 100 cores, we could measure $\Omega$ for 6 crashed neurons in roughly 4 days, but $\Omega$ for 7 crashed neurons would already take a month, and $\Omega$ for 8 crashed neurons is out of reach with more than 4 months of execution.

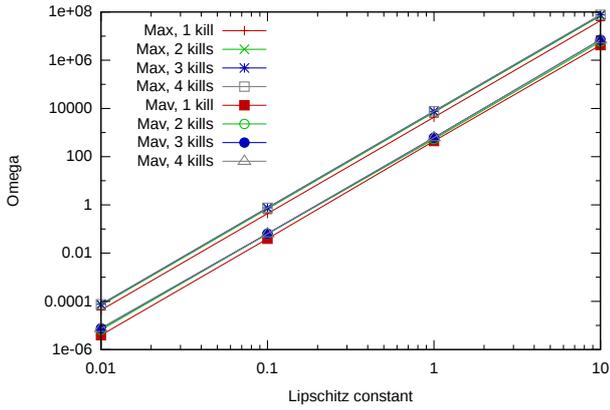

Fig. 3. Average error ($\Omega_{mav}$ and $\Omega_{max}$) for different number of crashed neurons on random networks, with linear rectifiers (ReLu), plotted as a function of the Lipschitz coefficient. The linear rectifiers act as pure propagator of the Lipschitz effect: we observe a clear polynomial dependency.

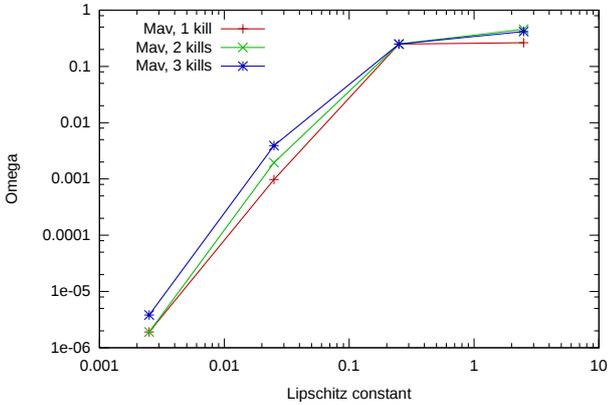

Fig. 4. Average error for different number of crashed neurons on random networks, with sigmoidal actival functions, plotted as a function of the Lipschitz coefficient. Sigmoid has a saturating effect that reduces the polynomial dependency on K when K increases (Cf. comments on tightness in the proof of Theorem 2.

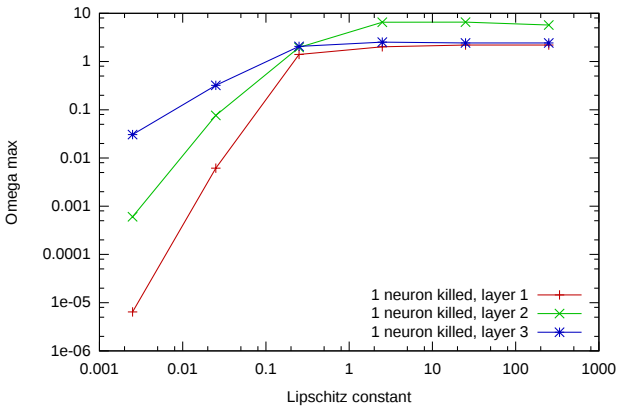

Fig. 5. Inversion of the depth effect when K crosses 1: average values of $\Omega$ on all crash situations of 1 neuron are plotted as a function of $K$, for three different depths: 1, 2 and 3. When K is smaller than 1, a deeper layer results in lower error, the opposite occurs when K becomes larger than 1

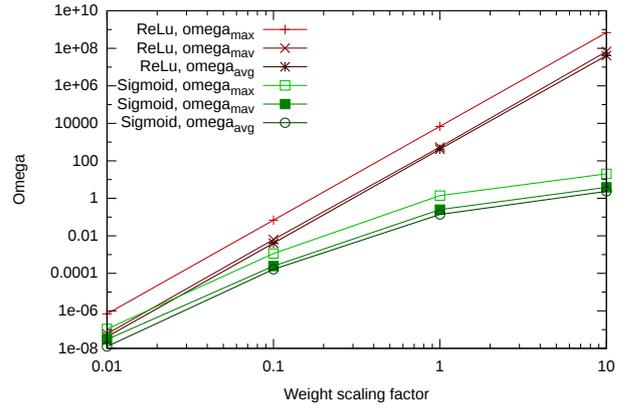

Fig. 6. Effect of scaling up or down the weights on the propagated error. Straight curves correspond to the case of ReLu as an activation function, while saturating behaviors correspond to sigmoid. The slope (in the case of the non saturating ReLu) is consistent with Equation 4: when all the weights are multiplied by 10, $Erf$ is multiplied by $10^{L+1}$.

### B. Role of the Weights

What happens to the error at the output when we change the scale of the weights of our random networks?

As we observe in Figure 6, for the ReLu activation function, scaling up the weights (and therefore the maximal weight involved in Equation 4) by a factor of 10, increases the value of Omega by $10^L$ where $L = 4$ is the number of layers of our network, which consistent of the multiplicative nature of $Erf$ suggested by Equation 4. In the case of sigmoid, we again obtain the late saturation when $K \geq 1$ similarly to what has been observed for the Lipschitz coefficient dependency.

### C. Orders of Magnitude on a Real-Life Case

Generating random networks is very informative to have a glimpse of the possible numerical phenomena occurring in neural networks and their altered versions when some of the neurons are crashed, however, if this was enough to provide experimental evidence for all the dependencies appearing in Equation 4, it does not provide orders of magnitude one could use in real life implementations.

"How many crashed neurons can my network tolerate if my margin of error is 0.15 averaged on all my inputs ?" - "I have trained three different networks with the same precision on a data-set (that I would not disclose), can you tell me which of the three will tolerate more crashes without testing them on my data-set ?". Those are the kind of questions Equation 4 tackles. In this part, we provide experimental evidence of how good does $Erf$ estimates (upper bounding) the propagated error $\Omega$.

Consider, for instance one of the most common use of neural networks: classifying handwritten digits, the final product is a neural network that, given an image **X**, will output a vector $F_{neu}(\mathbf{X}) \in (0, 1)^{10}$ such that the $i^{th}$ coordinate of $F_{neu}(\mathbf{X})$ is the probability that **X** is a handwritten version of the digit $i$. Using the MNIST database, we trained several neural networks to perform this digits classification task, we could achieve precision levels on a test set (of data unseen during training) ranging from 65% to 97% depending on the network topology, the activation function and the rigor of our robustness constraints. In particular, we used the dropout procedure [27], [33] consisting, as the name suggests, in dropping out neurons during training. We varied the probability with which a neuron is dropped out, from 0 to 0.3 by steps of 0.1.

In Figure 7, we observe that the networks learning with the dropout procedure are more robust, in the sense that their overall $\Omega$ is lower than the network for which the dropout rate was 0. Dropout acts as a robustness enhancer, and the less dropout is performed during learning, the closer is the measured robustness to the upper bound of Theorem 2. It is also worth noticing that the theoretical prediction $Erf$ is a good indicator of the range of $\Omega$, in that sense, looking at $Erf$ alone is enough to predict that the upper network cannot tolerate 3 crashes if the requirement is that the average error on the output does not exceed 0.15 (i.e that the network predicts the correct probability for a digit $i$ to be in image **X** by an average of 85%) while the three lower networks can tolerate 3 crashes and keep that requirement. Similarly, only the two later networks (dropouts of 0.20 and 0.30) will tolerate 1 crash failure if a more severe requirement of keeping a maximum error of 0.03. Note also that higher dropouts during learning make $Erf$ a safer estimation for robustness (the real error $\Omega$ is much lower than the predicted one $Erf$). This is consistent with the general intuition that networks trained while dropping out neurons will learn to perform a task while not always relying on "every neuron on-board" [27]. Finally, in Figure 10, we observe that even in the least robust situation (no dropout in the learning phase), the measured error is still safely estimated by the theoretical estimator as shown by the standard deviation.

## V. ANALYSIS

### A. Cost of Robustness

Obviously, building robust neural networks never comes without a cost. Indeed it can be tempting from the observations of the effect of the Lipschitz coefficient to choose a low one in order to increase robustness. Unfortunately, one cannot just choose to decrease this coefficient arbitrarily without consequences on the cost of learning. A naive perspective after looking at the activation function profile (Figure 2) is to consider that the higher the Lipschitz coefisscient -i.e the steeper is the slope in the transition zone (Figure 2), the most discriminating is the activation function, and therefore the easier it is for a network to be trained to classify a given data-set. Maybe surprisingly, our experiments contradict this intuition. The number of epochs needed to converge to a given precision target can be so random (depending on how the weights were initialized for example) that for 100 runs over 9 different values of K, we observe a mean number of needed epochs so random that it is in the same range as its standard deviation (Figure 8).We also observe, as discussed sometimes in the literature, that for a given task, there always is some sweet spot for K [40] where learning is the most efficient.

The bad news for designers whose goal is to build a robust network is that they have to give up on learning efficiency in order to gain robustness: decreasing K to improve robustness can lead you further from that sweet spot. But the reward in robustness is worth the sacrifice as shown in figures 3 and 4.

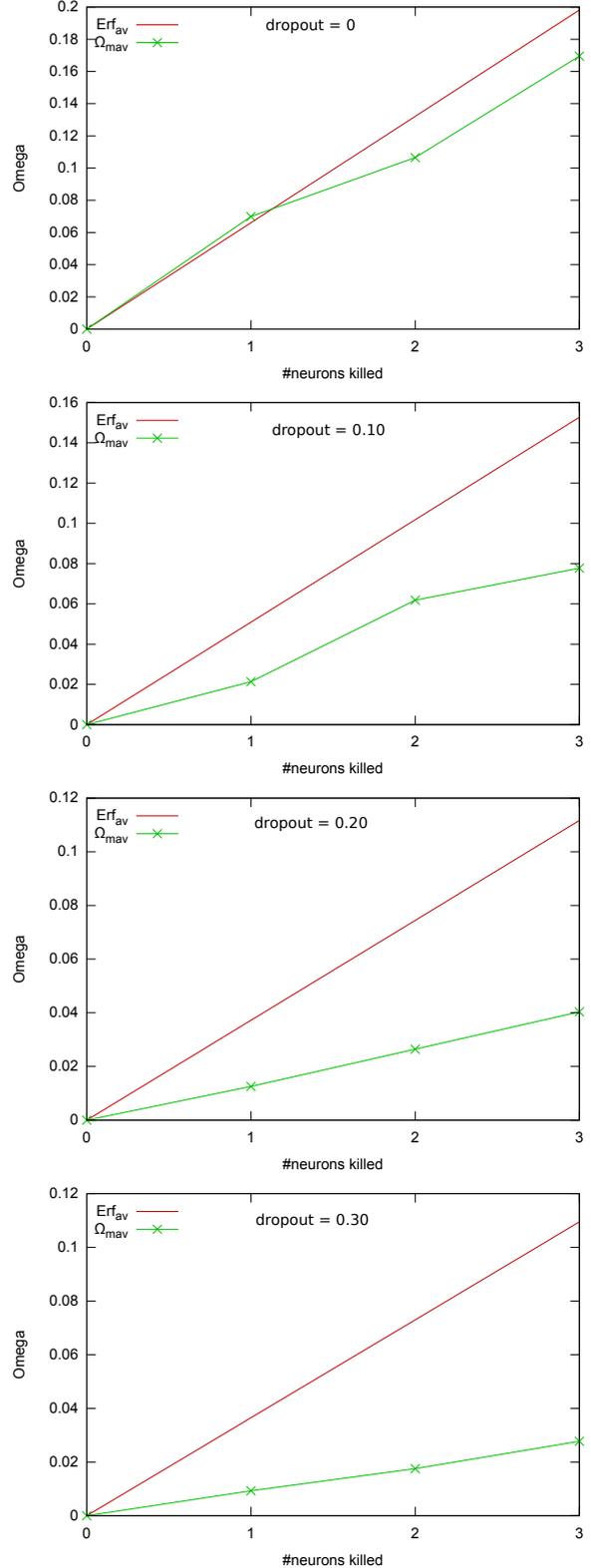

Fig. 7. Quality of $Erf$ as an estimator of $\Omega$: $Erf_{av}$ as an estimator of the observed $\Omega_{mav}$. The dropout rate is respectively 0, 0.1, 0.2 and 0.3 from the higher plot to the lower one. The more dropout is used in learning, the more robust is the network compared to the theoretical upper bound.

Figure 9 gives a rough schematic view of the dilemma we faced when trying to combine robustness (low value of the Lipschitz coefficient K), and reasonable convergence time to train a network to reach a given precision level.

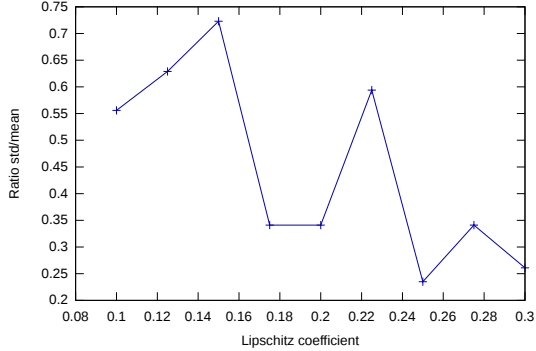

Fig. 8. The ratio of the standard deviation to the mean value of the learning time for 100 runs at each value of the Lipschitz coefficient ranging from 0.1 to 0.3.

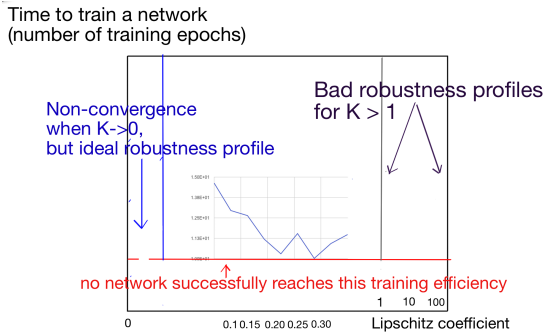

Fig. 9. Schematic view of the robustness-ease of training dilemma. Bellow the red line is the regime for which no 3-layers network of 14 neurons each could reach 95% accuracy on MNIST. On the left of the blue line, is an utopic robustness zone where K is close to zero but where no convergence is observed after a long number of epochs (»1000). The regime on the right of the magenta vertical line (K>1), is of no robustness interest since the output error explodes as shown by Figure 3 and Figure 4 and predicted by the $K^l$ theoretical dependency.

### B. Realism versus Safety

The second take-away from our experiments is that, if the worst-case bound on the error provides a tight bound defining the robustness criteria, leaving no doubt that the network will be correct (inside the margin enabled by the over-provision), this criteria reveals to be too harsh when we want the network to behave correctly withing reasonable average case situations. In those situations, the estimator $Erf_{av}$ is closer to the expected error (cf. comments on equality cases in the proof of Theorem 1).

In both cases, a designer can submit his network weights to our implementation [37], and **without having to disclose his training set, or specifying the task performed by his network**, get estimators of how much error his network will generate, both on average and worst case, given a desired budget of failures.

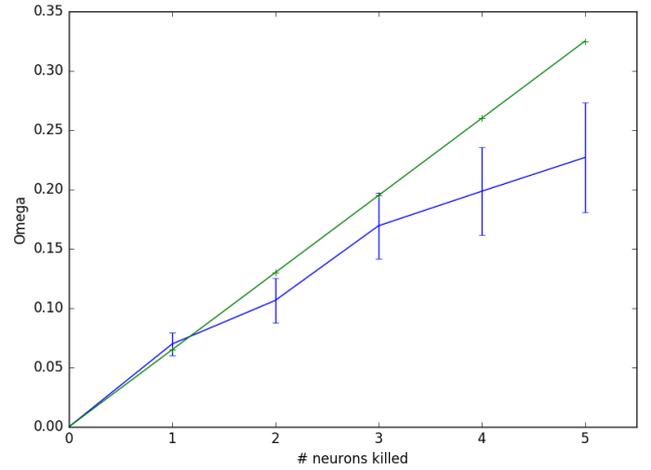

Fig. 10. The average measured error $Omega_{av}$ (blue) exhibits a close behaviour to the theoretical predication $Erf_{av}$ as showed by the measured standard deviation (vertical bars).

## VI. CONCLUSION

Before the actual regain of interest in deep learning, the interest in neural networks faded in the mid-nineties, so did the interest in their tolerance to the failure of individual neurons. With the growth in modern neuromorphic hardware, the question of robustness to individual neurons failure becomes critical. This paper is a step towards a better understanding of this robustness, while most of the recent work on robustness has been driven by the coarse-grained view where the unit of failure is a machine, not a single neuron [20], [21], [22].

In fact, moving from the traditional view, where the unit of failure is a machine, hosting the neural network as a software abstraction, to a more granular view where a single neuron is the unit of failure, is somewhat equivalent to the approach taken in a paper by Borkar [41] on the failures of single transistors, or the paper by Constantinescu on VLSI circuits [42]. In all of these cases, the unit of failure is not a whole machine, but just a microscopic part of it (a neuron, a transistor, etc).

Our paper shows that the robustness of a neural network can be assessed solely in terms of constants of the networks, without the need to analyze the numerous inputs such a network can deal with during its lifetime. Or to test the shut down of every possible subset of neurons of a given size.

More precisely, we provided theoretical proofs and experimental evidence that the error propagated to the output due to neurons failures is a function that grows, even on average situations, polynomially with K, the Lipschitz coefficient of the activation function. This polynomial dependency on K has a degree equal to the depth of the layer where the failure happens (the error has therefore an exponential dependency on the depth). We also showed that the propagated error to the output is proportional to the product, over all the layers of the network, of the average absolute values of weights per layer. This is yet another exponential dependency on the depth that adds up to the dependency created by the Lipschitz factor. Based on those observations, we conclude that neural neutral networks with steeper activation functions (higher Lipschitz coefficient) are less robust compared to their counterparts with gentle activation functions. The catch being that robustness comes at the cost of learning: for a neural network to learn, the Lipschitz coefficient should be high enough, however, theoretically assessing how high this coefficient should be, remains a challenging open problem [40], [43].

**Acknowledgment.** This work is supported by the Swiss National Foundation under the grant *A Theoretical Approach to Robustness in Biological Algorithms* 200021 169588 TARBDA and by the European ERC Grant 339539 - AOC.